\documentclass[conference]{IEEEtran}
\IEEEoverridecommandlockouts
\usepackage{cite}
\usepackage{amsmath,amssymb,amsfonts}
\usepackage{algorithmic}
\usepackage{graphicx}
\usepackage{textcomp}
\usepackage{xcolor}
\usepackage{hyperref}
\usepackage{subcaption}
\usepackage{fancyhdr}
\usepackage{stfloats} 
\usepackage{tabularx}
\usepackage{subcaption}
\usepackage{fancyhdr}
\usepackage{booktabs}
\usepackage{multirow}
\usepackage{siunitx}
\usepackage{pdfpages}
\usepackage{url}
\usepackage{mwe, multicol}
\usepackage{float}
\usepackage{tikz}
\usepackage{lipsum}

\def\BibTeX{{\rm B\kern-.05em{\sc i\kern-.025em b}\kern-.08em
    T\kern-.1667em\lower.7ex\hbox{E}\kern-.125emX}}

\newcommand\copyrighttext{%
  \footnotesize \textcopyright 2023 IEEE. Personal use of this material is permitted. Permission from IEEE must be obtained for all other uses, in any current or future media, including reprinting/republishing this material for advertising or promotional purposes, creating new collective works, for resale or redistribution to servers or lists, or reuse of any copyrighted component of this work in other works.}
\newcommand\copyrightnotice{%
\begin{tikzpicture}[remember picture,overlay]
\node[anchor=south,yshift=10pt] at (current page.south) {\fbox{\parbox{\dimexpr\textwidth-\fboxsep-\fboxrule\relax}{\copyrighttext}}};
\end{tikzpicture}%
}

\makeatletter

\let\old@ps@IEEEtitlepagestyle\ps@IEEEtitlepagestyle
\def\confheader#1{%
    \def\ps@IEEEtitlepagestyle{%
        \old@ps@IEEEtitlepagestyle%
        \def\@oddhead{\strut\hfill#1\hfill\strut}%
        \def\@evenhead{\strut\hfill#1\hfill\strut}%
    }%
    \ps@headings%
}
\makeatother

\confheader{%
    \parbox{50cm}{\textbf{Accepted and presented at "2023 26th International Conference on\\ Computer and Information Technology (ICCIT) 13-15 December 2023, Cox’s Bazar, Bangladesh"}}
}

\begin{document}

\title{Unmasking Deepfake Faces from Videos Using An Explainable Cost-Sensitive Deep Learning Approach \\

}

\author{\IEEEauthorblockN{Faysal Mahmud\IEEEauthorrefmark{1}, Yusha Abdullah\IEEEauthorrefmark{2}, Minhajul Islam\IEEEauthorrefmark{3}, Tahsin Aziz\IEEEauthorrefmark{4}}
\IEEEauthorblockA{Department of Computer Science and Engineering\\
Ahsanullah University of Science and Technology, Dhaka, Bangladesh\\
\{faysalmmud\IEEEauthorrefmark{1}, yusha.abd\IEEEauthorrefmark{2}, minhajulislam.aust\IEEEauthorrefmark{3}, tahsinaziz.cse\IEEEauthorrefmark{4}\}@gmail.com}
}

\maketitle
\copyrightnotice


\begin{abstract}
Deepfake technology is widely used, which has led to serious worries about the authenticity of digital media, making the need for trustworthy deepfake face recognition techniques more urgent than ever. This study employs a resource-effective and transparent cost-sensitive deep learning method to effectively detect deepfake faces in videos. To create a reliable deepfake detection system, four pre-trained Convolutional Neural Network (CNN) models: XceptionNet, InceptionResNetV2, EfficientNetV2S, and EfficientNetV2M were used. FaceForensics++ and CelebDf-V2 as benchmark datasets were used to assess the performance of our method. To efficiently process video data, key frame extraction was used as a feature extraction technique. Our main contribution is to show the model's adaptability and effectiveness in correctly identifying deepfake faces in videos. Furthermore, a cost-sensitive neural network method was applied to solve the dataset imbalance issue that arises frequently in deepfake detection. The XceptionNet model on the CelebDf-V2 dataset gave the proposed methodology a 98\% accuracy, which was the highest possible whereas, the InceptionResNetV2 model, achieves an accuracy of 94\% on the FaceForensics++ dataset. Source Code: \href{https://github.com/Faysal-MD/Unmasking-Deepfake-Faces-from-Videos-An-Explainable-Cost-Sensitive-Deep-Learning-Approach-IEEE2023}{https://github.com/Faysal-MD/Unmasking-Deepfake-Faces-from-Videos-An-Explainable-Cost-Sensitive-Deep-Learning-Approach-IEEE2023}

\end{abstract}



\section{Introduction}
Deepfake combines deep learning and fake technology. The deep learning capability of artificial intelligence allows for the detection and creation of deepfakes \cite{wubet2020deepfake}. Generative adversarial networks (GAN) produce deepfakes, which have two existing machine-learning models. The deepfakes are produced by one model after it has been trained on a dataset and detected by another model. Deepfakes produce fake videos, news, photographs, and terrorist incidents. It might cause misinformation circulating online platforms, even if it sounds like an exciting approach to create false images and videos of anything or certain people. Spreading bogus videos and images on social media leads people to disbelieve the truth. Deepfakes are progressively having an impact on security, people, groups, businesses, religions, and democracies.
\par
Deepfake video detection tasks have very limited dataset availability and are often imbalanced. A pipeline to solve the problem of data imbalance and extract key frames from videos has been proposed in this study. The main idea behind key frame extraction is to reduce the volume of data and computation time in a video while maintaining its essential content and context, whereas cost-sensitive methods are used in situations where the goal is to give more importance or bias toward classes that may have a higher cost associated with misclassification. Several approaches use key frame extraction \cite{mitra2021machine} and cost-sensitive neural networks \cite{tanvir2023explainable}, \cite{zhou2005training} for various detection tasks. In this study, various pre-trained CNN-based models were assessed to identify deepfake faces in videos.
\par 
There is a significant risk involved as we do not understand how these sophisticated neural networks generate their predictions. Users undoubtedly rely on the models as they hardly try to understand what is happening in the model's backend. The ability to understand how a model produces any kind of decision would be great. To analyze the weights and inner workings of a neural network as well as to explain the output of the pre-trained models used in the proposed methodology, we used a variety of Gradient-based Explainable AI tools, including SmoothGrad, GradCAM, GradCAM++, and Faster Score-CAM. The models in this study were also trained using two separate imbalanced datasets. To sum up, we have added the following to this research: 

\begin{itemize}
    \item The key frame extraction approach was employed to reduce the amount of data and processing time in a video while keeping its important content.
    \item To handle imbalanced datasets, we have suggested a method for deepfake face detection from videos. 
    \item For an imbalanced dataset, we implemented the idea of a cost-sensitive neural network.
    \item The prediction of the trained models is explained by the Explainable AI techniques (XAI).
\end{itemize}


\section{Literature Review}
Lee and Kim \cite{lee2021deepfake} came up with an idea to extract the rate of change of adjacent frames to check whether a video is manipulated or not. Faceforensics++ \cite{rossler2019faceforensics++} and DFDC \cite{dolhansky2020deepfake} datasets were used in this study where three hundred frames were extracted from each video. Their proposed Deep Neural Network architecture provided a higher accuracy of 97\% than any other existing methods. However, the frame image's metric had been altered showing their method, which can lead to ineffective detection of good-quality datasets. They are currently working on this matter.
For Deepfake Video Detection, Xu et al. \cite{xu2021deepfake} proposed a method for constructing texture features and processing them with a feature selection method. This discriminant feature vector is then passed on to SVM for classification. DeepFake-TIMIT \cite{korshunov2018deepfakes}, Celeb-DF \cite{li2019celeb}, FaceForensics++ \cite{rossler2019faceforensics++}, and DFDC \cite{dolhansky2020deepfake} datasets were used in this experiment. They have achieved the highest accuracy of 91.2\% on C40 quality videos of the FaceForensics++ dataset. Though the texture details of the deepfake videos are insufficient, they think this method can work well if new deepfake videos with defective texture features are released in the future.
Two-dimensional global discrete Cosine transforms (2D-GDCT) are used in the method that Kohli et al. \cite{kohli2021detecting} suggested to extract faces from a target video and convert them into the frequency domain. They used FaceForensics++ \cite{rossler2019faceforensics++} and CelebDf-V2 \cite{li2019celeb} datasets in their study. Then, to identify fake facial images, a 3-layered frequency convolutional neural network (CNN) is used. At first, they split the target video into frames and further converted it into frequency, and then a convolutional neural network was trained to learn the facial image's frequency features. The highest accuracy from their proposed model is 86.08\% and for average and maximum pooling technique is 85.24\%. Their proposed model performed better in terms of detecting pristine faces. But in total accuracy, XceptionNet performed better. 
Kim et al. \cite{kim2021fretal} introduced two models on the FaceForensics++ dataset \cite{rossler2019faceforensics++}. The first one trained with XceptionNet and the second one needed weights of the first model to train. Then they used Feature-based Representation Learning to find out same feature in different deepfake videos. They marked it as feature transfer learning. XceptionNet is the backbone of their proposed model FReTAL. Their proposed model scores 86.97\% accuracy in deepfake video.


\section{Background Study}
\subsection{XceptionNet}
François Chollet introduced XceptionNet \cite{Chollet_2017_CVPR} in 2017, a ground-breaking convolutional neural network (CNN) that will transform how images are categorized. By utilizing depthwise separable convolutions, this cutting-edge architecture drastically lowers the number of parameters and computing requirements. Its extraordinary efficiency is a result of the inclusion of crucial features including global average pooling, batch normalization, skip connections, and ReLU activations. XceptionNet can also use dropout layers to avoid overfitting. This network, which is renowned for performing exceptionally well in image classification tasks, strikes the optimum mix between complexity and efficiency, making it the best option for a variety of deep learning applications.

\subsection{EfficientNetV2S}
EfficientNetV2S is a highly efficient convolutional neural network (CNN) architecture that aims to be both parameter and FLOP (floating-point operation) efficient. The EfficientNetV2 architecture, created by Google AI \cite{tan2021efficientnetv2} in 2021, has been scaled down and made faster. Comprising MBConv and Fused-MBConv blocks, it achieves state-of-the-art results on tasks like ImageNet classification with just 22.2 million parameters and 3.9 billion FLOPs.It has been demonstrated that EfficientNetV2S outperforms other state-of-the-art models in terms of performance on a range of image classification tasks. Especially in situations where efficiency is a key requirement, such as mobile devices and embedded systems.

\subsection{EfficientNetV2M}
EfficientNetV2M is a highly efficient and powerful convolutional neural network(CNN) architecture designed by Google AI \cite{tan2021efficientnetv2} for image classification. It features inverted residual blocks, mobile inverted bottleneck convolutions (MBConv blocks), fused mobile inverted bottleneck convolutions (Fused-MBConv blocks), and squeeze-and-excitation (SE) modules to optimize both accuracy and computational efficiency. The architecture is organized into stages with varying numbers of blocks, allowing for flexibility in model complexity. A fully connected layer and Global average pooling produce final predictions. Pre-trained on ImageNet, EfficientNetV2M is ideal for tasks like image recognition, object detection, and image segmentation, making it suitable for transfer learning.

\subsection{InceptionResNetV2}
InceptionResNetV2 is a convolutional neural network that fuses the strengths of Inception and ResNet architectures \cite{szegedy2016inceptionv4}. It combines dimension-reduction blocks, skips connections to address gradient vanishing problems, and Inception blocks to capture multi-scale characteristics. Global average pooling, fully connected layers, and a softmax output are used after feature extraction. It is well-known for its great accuracy and processing efficiency, making it a preferred option for many computer vision jobs. While the number of layers may vary, the architecture it normally employs is deep, demonstrating the versatility and efficiency of the system in complicated picture processing.

\subsection{Explainable AI}
The activities in order to increase human comprehension are the main objective of an explainable AI (XAI) \cite{gunning2019xai}. The XAI system should be able to describe its capabilities and comprehension as well as what it has accomplished and is doing at the moment, along with what is coming next. It should be able to provide important details about itself based on its behaviors. 


\section{Dataset}
There are a few publicly available databases that include both real and fake videos. The FaceForensics++ \cite{rossler2019faceforensics++} dataset and Celeb-DF \cite{li2019celeb} were discovered the most frequently used sources of data, based on our research into previous studies. The frequency of the datasets is listed below in Table \ref{tab:dataset}

\begin{table}[h]
    \caption{Information of Datasets}
    \label{tab:dataset}
    \centering
    \begin{tabular}{|l|c|c|c|c|}
        \hline
        \textbf{Name}     & \textbf{Real} & \textbf{Fake} & \textbf{Forgery Methods} \\
        \hline
        FaceForensics++   & 1000          & 4000          & 4             \\ \hline
        Celeb-DF          & 590           & 5639          & 1           \\ \hline
    \end{tabular}
\end{table}


\section{Methodology}
\begin{figure}[h]
    \centerline{\includegraphics[width = 0.45\textwidth]{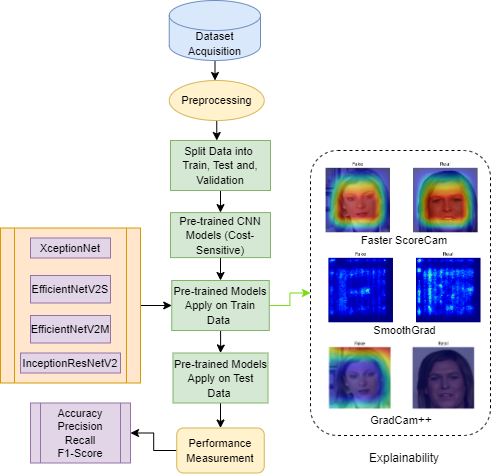}}
    \caption{Proposed Methodology}
    \label{fig:methodology}
\end{figure}

The proposed methodology is demonstrated in this section. Figure \ref{fig:methodology} displays the complete procedure of the methodology.

\subsection{Data Preprocessing}
\begin{figure}[h]
    \centerline{\includegraphics[width = 0.45\textwidth]{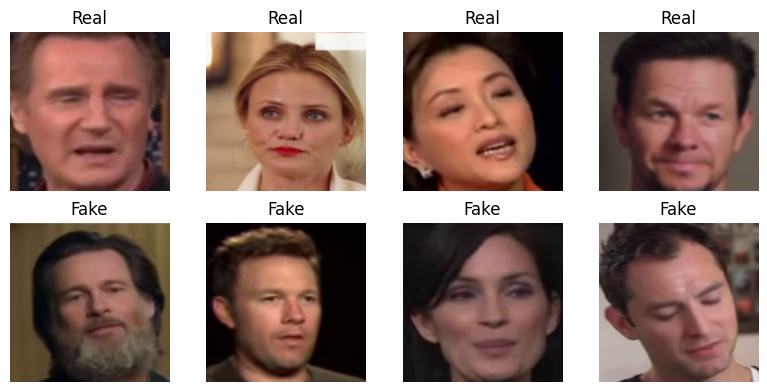}}
    \caption{Preprocessed Data}
    \label{fig:PreprocessedData}
\end{figure}

Firstly, it was determined that if the video was corrupted or not. The video file is immediately erased if it is discovered to be corrupted. Next, the Python "face recognition" package was used to extract faces from videos to find faces inside the video. It effectively filters out unnecessary background elements to extract just the faces. Each frame was resized to a constant 224$\times$224 resolution during this process, and the video has 30 frames per second (fps) frame rate. The final dataset visualization is presented in Figure \ref{fig:PreprocessedData}

The proposed approach used inter-frame differences during the key frame extraction step. The basic idea is simple: after loading the video, how much each pair of frames differs from one another was figured out. Then it was put in a local maximum detection process. Keyframes are specifically those for which the average inter-frame difference is at the local maximum. It is important to notice that noise was removed effectively and stopped the repeated extraction of frames from similar situations by smoothing the average difference values before completing the local maxima computation. Ultimately, the raw image data was transformed into a numpy array to further optimize computation time and improve training quality.

\subsection{Data Split Ratio}
Training (80\%), Testing (10\%), and Validation (10\%) were the three divisions of the dataset. To improve the distribution of data before distribution, the entire dataset was stratified. Dataset information is shown in Table \ref{tab:celebdf} and Table \ref{tab:ff++}.

\begin{table}[h]
    \caption{Data distribution of CelebDf-V2}
    \label{tab:celebdf}
    \centering
    \begin{tabular}{|l|c|c|c|c|}
        \hline
        \textbf{Class} & \textbf{Train} & \textbf{Validation} & \textbf{Test} \\
        \hline
        Fake           & 8995           & 1000                & 1111          \\ \hline
        Real           & 2383           & 265                 & 294           \\ \hline
    \end{tabular}
\end{table}

\begin{table}[h]
    \caption{Data distribution of FaceForensics++}
    \label{tab:ff++}
    \centering
    \begin{tabular}{|l|c|c|c|c|}
        \hline
        \textbf{Class} & \textbf{Train} & \textbf{Validation} & \textbf{Test} \\
        \hline
        Fake           & 6728           & 748                & 831          \\ \hline
        Real           & 3847           & 428                & 475           \\ \hline
    \end{tabular}
\end{table}

\subsection{Cost-sensitive}
To fix the class imbalance in the training data, class weights are first generated to give minority classes greater weight during model training. Based on the distribution of class labels in the training data, these class weights are calculated. The next step is to compile these class weights into a dictionary for use in model training. This will allow a more balanced learning process.

\subsection{Model Trainig}
In this study, we implemented four pre-trained CNN models (XceptionNet, EfficientNetV2S, EfficientNetV2M, and InceptionResNetV2). A 0.001 learning rate was applied for all the models. If the model consistently performs poorly for a set number of epochs, the ReduceLROnPlateau class was used to lower the learning rate. In each model, there is a batch size of 16 and the optimization technique has been implemented using the Adam optimizer. The output of the base model was enhanced with the GlobalAveragePooling2D layer, followed by the ReLU activation function and a Dense layer. A dropout layer with a rate of 0.5 is employed to prevent overfitting. At the final dense layer of each model, the Softmax activation function was employed. 
 
\section{Experimental Result}
\subsection{Result Analysis}
\begin{figure}[h]
  \centering
  \subfloat[XceptionNet]{\includegraphics[width=4.3cm]{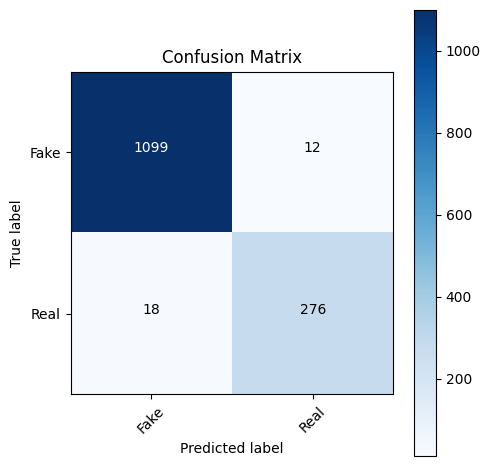}\label{fig:image1}}
  \hfil
  \subfloat[InceptionResNetV2]{\includegraphics[width=4.3cm]{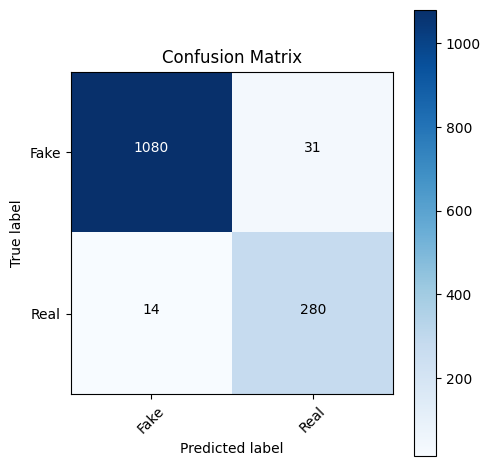}\label{fig:image3}}
  \hfil
  \subfloat[EfficientNetV2S]{\includegraphics[width=4.3cm]{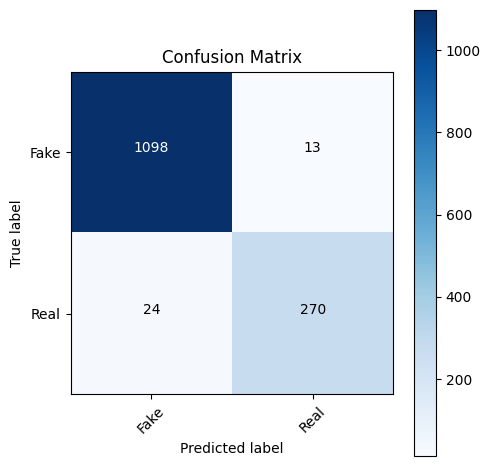}\label{fig:image2}}
  \hfil
  \subfloat[EfficientNetV2M]{\includegraphics[width=4.3cm]{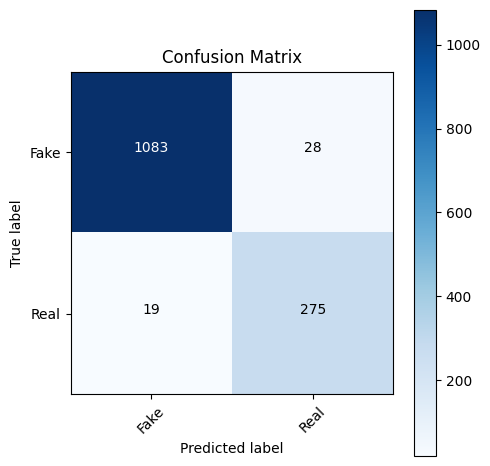}\label{fig:image4}}
  \caption{Confusion Matrix of All Pre-trained Models Based on CelebDF-V2 Dataset}
  \label{fig:cm-celeb}
\end{figure}

\begin{figure}[h]
  \centering
  \subfloat[Original Image]{\includegraphics[width=4cm]{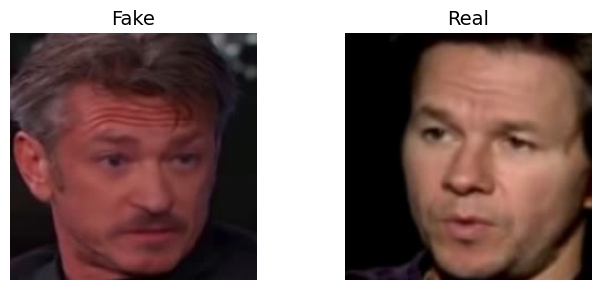}\label{fig:image1}}
  \hfil
  \subfloat[SmoothGrad]{\includegraphics[width=4cm]{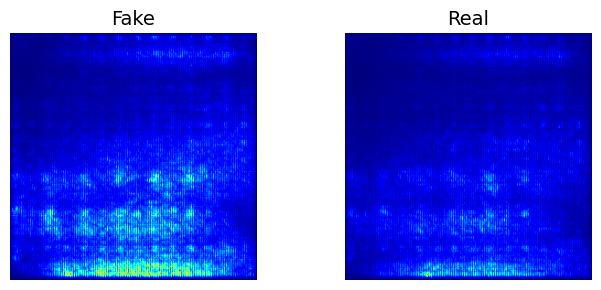}\label{fig:image1}}
  \hfil
  \subfloat[GradCam]{\includegraphics[width=4cm]{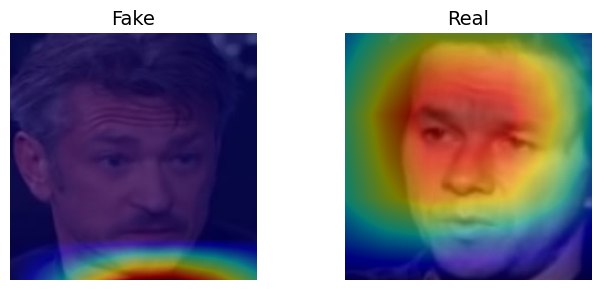}\label{fig:image3}}
  \hfil
  \subfloat[Faster ScoreCam]{\includegraphics[width=4cm]{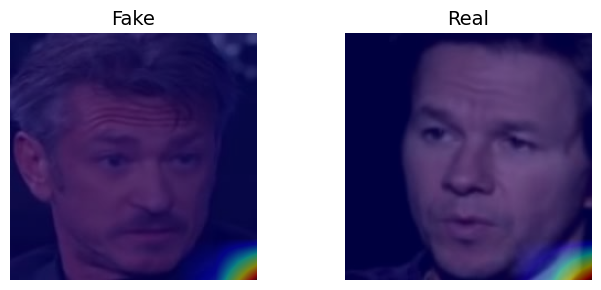}\label{fig:image4}}
  \caption{Visualization of sample outputs how well XceptionNet model can capture the faces on CelebDf-V2 dataset}
  \label{fig:xai-xn-celeb}
\end{figure}

\begin{figure}[h]
  \centering
  \subfloat[Original Image]{\includegraphics[width=4cm]{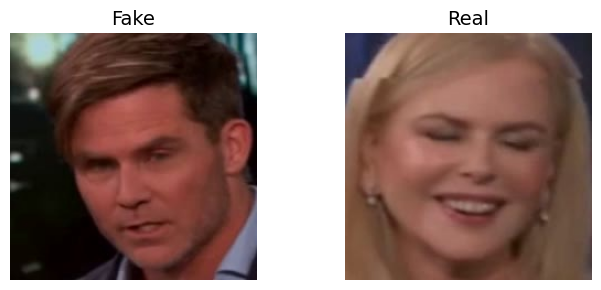}\label{fig:image1}}
  \hfil
  \subfloat[SmoothGrad]{\includegraphics[width=4cm]{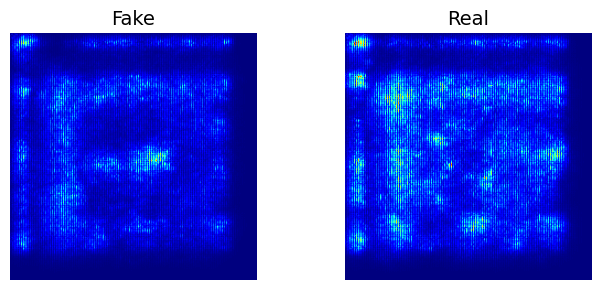}\label{fig:image1}}
  \hfil
  \subfloat[GradCam]{\includegraphics[width=4cm]{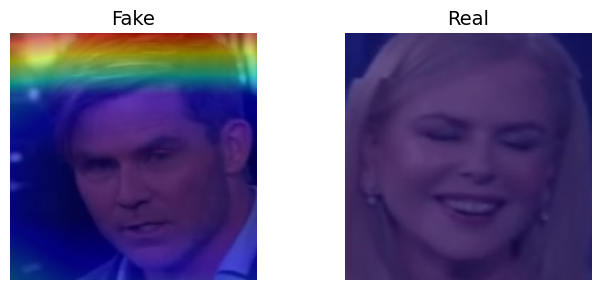}\label{fig:image3}}
  \hfil
  \subfloat[GradCam++]{\includegraphics[width=4cm]{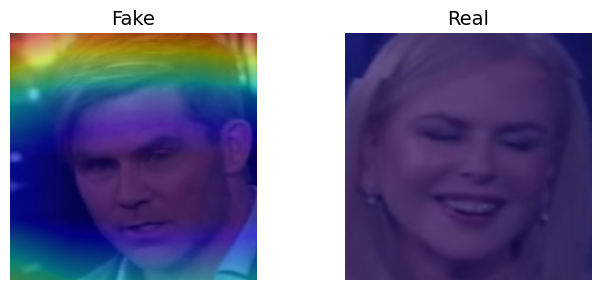}\label{fig:image2}}
  \caption{Visualization of sample outputs how well InceptionResNetV2 model can capture the faces on CelebDf-V2 dataset}
  \label{fig:xai-irnv2-celeb}
\end{figure}

\begin{figure}[h]
    \centering
    \subfloat[Original Image]{\includegraphics[width=4cm]{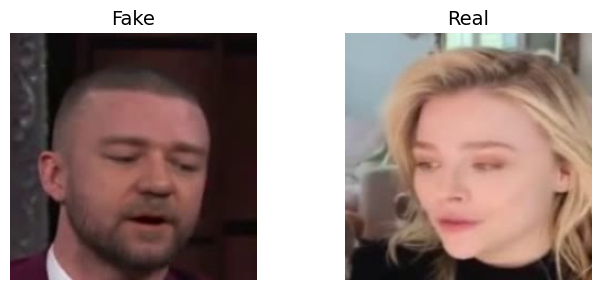}\label{fig:image1}}
    \hfil
    \subfloat[Faster ScoreCam]{\includegraphics[width=4cm]{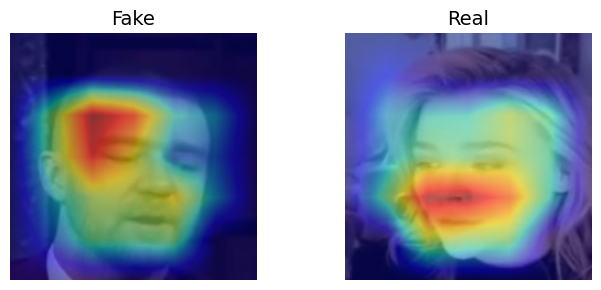}\label{fig:image1}}
    \hfil
    \subfloat[Original Image]{\includegraphics[width=4cm]{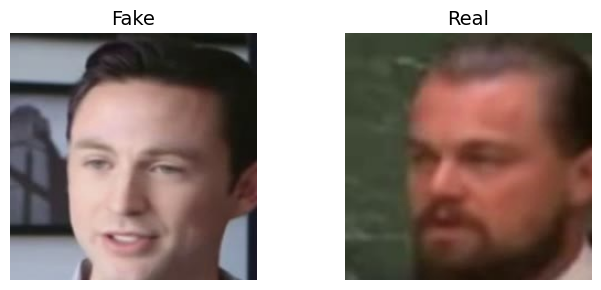}\label{fig:image1}}
    \hfil
    \subfloat[Faster ScoreCam]{\includegraphics[width=4cm]{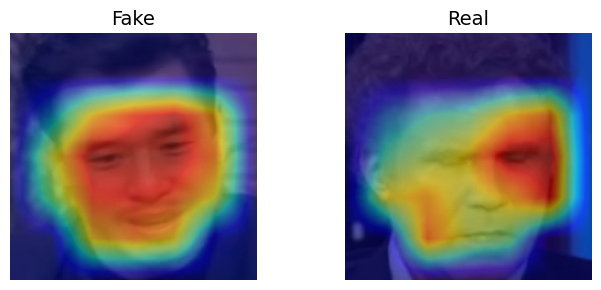}\label{fig:image1}}
    \caption{Visualization of sample outputs how well EfficientNetV2S and EfficientNetV2M model can capture the faces on CelebDf-V2 dataset: First row represents the EfficientNetV2S and second row represents EfficientNetV2S model}
    \label{fig:xai-env2-celeb}
\end{figure}

\begin{figure}[htbp]
  \centering
  \subfloat[XceptionNet]{\includegraphics[width=4cm]{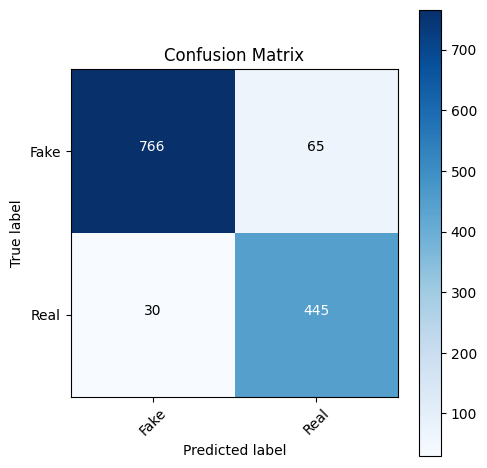}\label{fig:image1}}
  \hfil
  \subfloat[InceptionResNetV2]{\includegraphics[width=4cm]{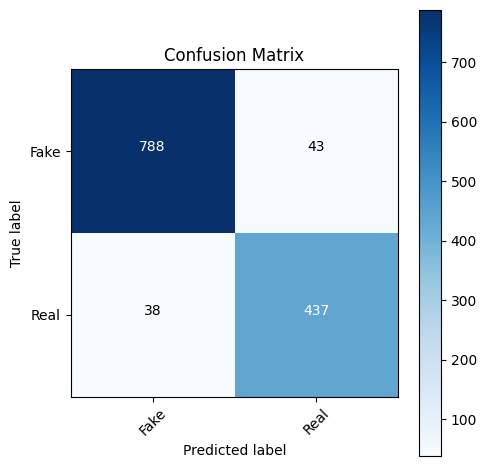}\label{fig:image3}}
  \hfil
  \subfloat[EfficientNetV2S]{\includegraphics[width=4cm]{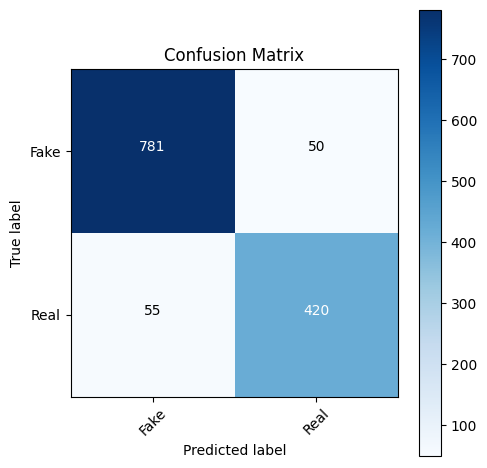}\label{fig:image2}}
  \hfil
  \subfloat[EfficientNetV2M]{\includegraphics[width=4cm]{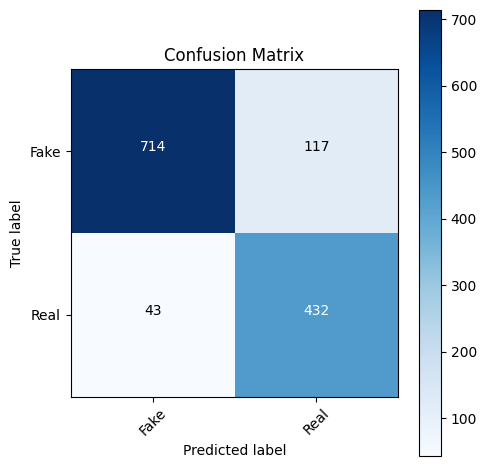}\label{fig:image4}}
  \caption{Confusion Matrix of All Pre-trained Models Based on FaceForensics++ Dataset}
  \label{fig:cm-ff}
\end{figure}

\begin{figure}[htbp]
  \centering
  \subfloat[Original Image]{\includegraphics[width=4cm]{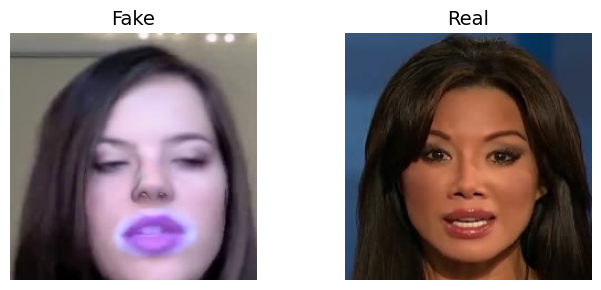}\label{fig:image1}}
  \hfil
  \subfloat[SmoothGrad]{\includegraphics[width=4cm]{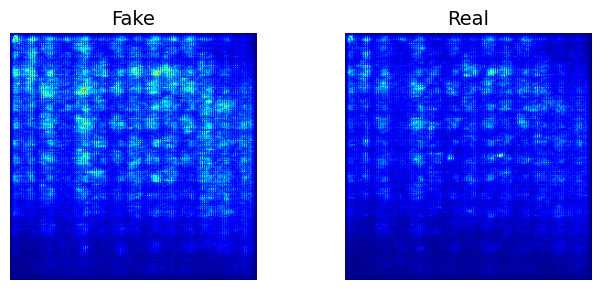}\label{fig:image1}}
  \hfil
  \subfloat[GradCam]{\includegraphics[width=4cm]{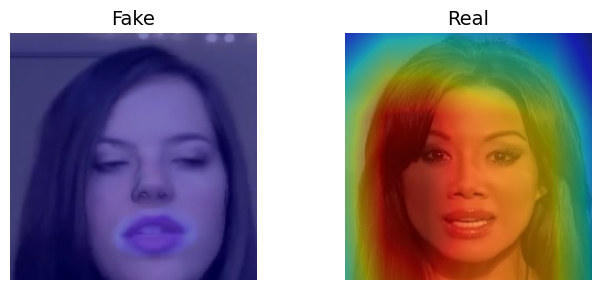}\label{fig:image3}}
  \hfil
  \subfloat[Faster ScoreCam]{\includegraphics[width=4cm]{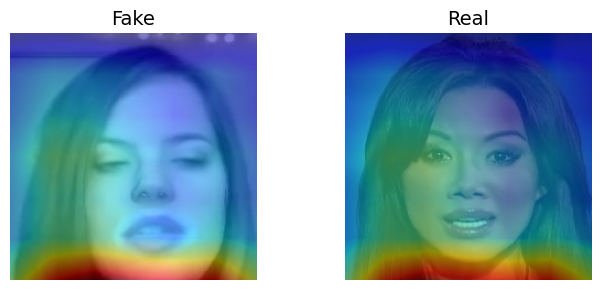}\label{fig:image4}}
  \caption{Visualization of sample outputs how well XceptionNet model can capture the faces on FaceForensics++ dataset}
  \label{fig:xai-xn-ff}
\end{figure}

\begin{figure}[htbp]
  \centering
  \subfloat[Original Image]{\includegraphics[width=4cm]{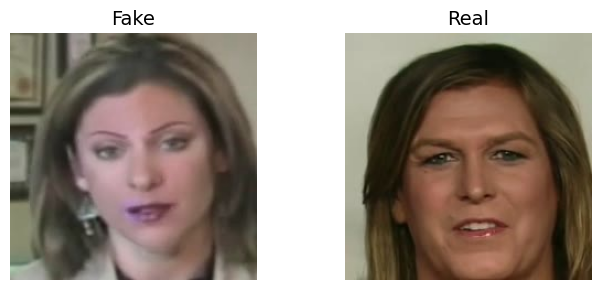}\label{fig:image1}}
  \hfil
  \subfloat[SmoothGrad]{\includegraphics[width=4cm]{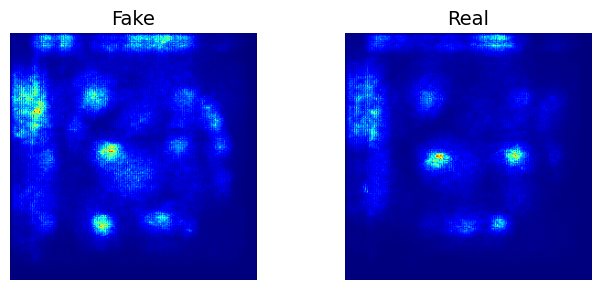}\label{fig:image1}}
  \hfil
  \subfloat[GradCam]{\includegraphics[width=4cm]{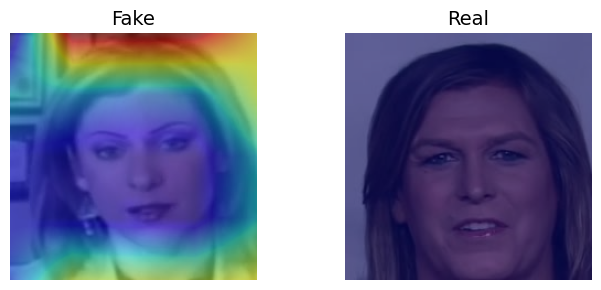}\label{fig:image3}}
  \hfil
  \subfloat[GradCam++]{\includegraphics[width=4cm]{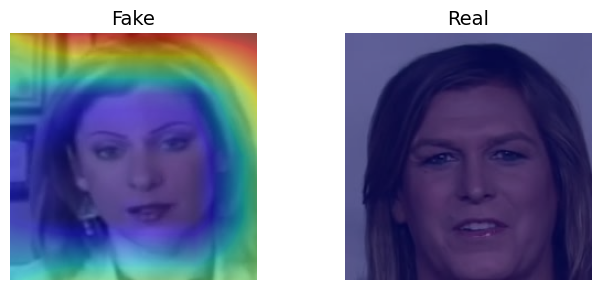}\label{fig:image2}}
  \caption{Visualization of sample outputs how well InceptionResNetV2 model can capture the faces on FaceForensics++ dataset}
  \label{fig:xai-irnv2-ff}
\end{figure}

\begin{figure}[htbp]
    \centering
    \subfloat[Original Image]{\includegraphics[width=4cm]{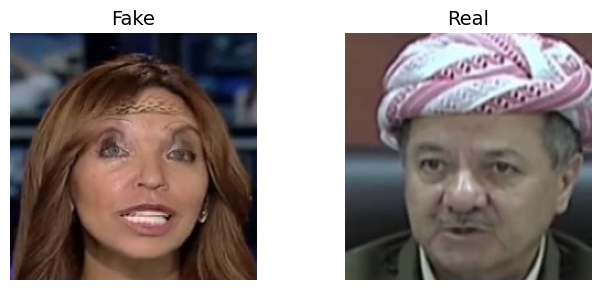}\label{fig:image1}}
    \hfil
    \subfloat[Faster ScoreCam]{\includegraphics[width=4cm]{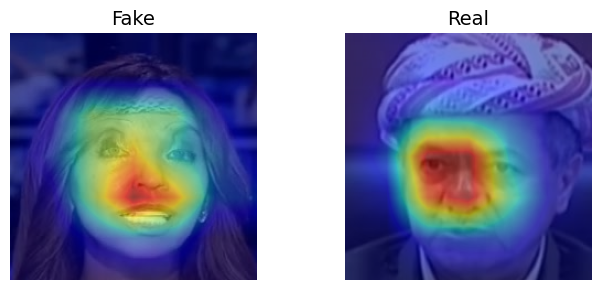}\label{fig:image1}}
    \hfil
    \subfloat[Original Image]{\includegraphics[width=4cm]{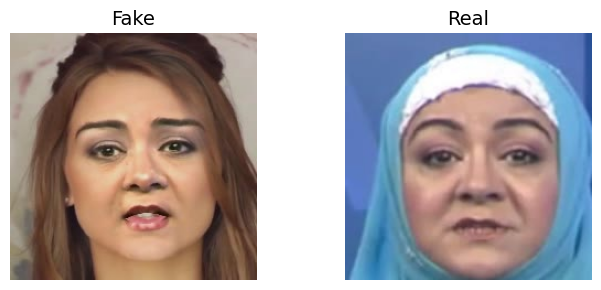}\label{fig:image1}}
    \hfil
    \subfloat[Faster ScoreCam]{\includegraphics[width=4cm]{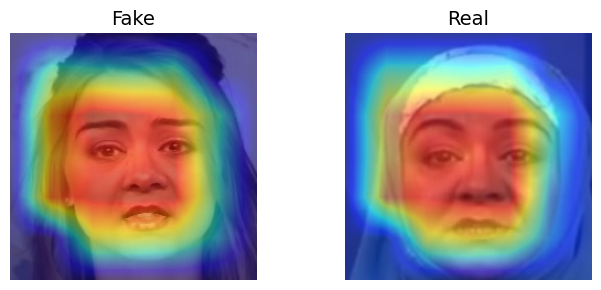}\label{fig:image1}}
    \caption{Visualization of sample outputs how well EfficientNetV2S and EfficientNetV2M model can capture the faces on FaceForensics++ dataset: First row represents the EfficientNetV2S and second row represents EfficientNetV2S model}
    \label{fig:xai-env2-ff}
\end{figure}

Two distinct datasets were used for all four pre-trained models. All of these models have mostly correctly predicted the fake faces from videos. The results obtained from the datasets using the four pre-trained models are presented in Table \ref{tab:pm-celeb} and \ref{tab:pm-ff++}. Moreover, the confusion matrix is shown in Figure \ref{fig:cm-celeb} for the CelebDf-V2 dataset and Figure \ref{fig:cm-ff} for the FaceForensics++ dataset, and the model's explainability is shown in Figure \ref{fig:xai-xn-celeb} - Figure \ref{fig:xai-env2-celeb} for CelebDF-V2 dataset and Figure \ref{fig:xai-xn-ff} - Figure \ref{fig:xai-env2-ff} for FaceForensics++ dataset.

\begin{table}[h]
    \caption{Performance Metrics of Weighted Average on CelebDf-V2 Dataset}
    \label{tab:pm-celeb}
    \centering
    \begin{tabular}{|l|c|c|c|c|}
        \hline
        \textbf{Model}     & \textbf{Accuracy} & \textbf{Precision} & \textbf{Recall} & \textbf{F1-Score} \\
        \hline
        XceptionNet        & \textbf{98\%}     & 0.98              & 0.98            & 0.98 \\ 
        \hline
        EfficientNetV2S    & 97\%              & 0.97              & 0.97            & 0.97 \\ 
        \hline
        EfficientNetV2M    & 97\%              & 0.97              & 0.97            & 0.97 \\ 
        \hline
        InceptionResNetV2  & 97\%              & 0.97              & 0.97            & 0.97 \\ 
        \hline
    \end{tabular}
\end{table}

Table \ref{tab:pm-celeb} shows that the models had great accuracy, with XceptionNet leading with 98\% and EfficientNetV2S, EfficientNetV2M, and InceptionResNetV2 following very closely with 97\% each. This suggests that the CelebDf-V2 Dataset performed exceptionally well overall.

Using Expainable AI it is determined how robust the model is. From Figure \ref{fig:xai-xn-celeb} to Figure \ref{fig:xai-env2-celeb} it has been proved that  Expainable AI produces different results for different models. In the case of CelebDF-V2 EfficientNetV2S and EfficientNetV2M have shown greater possibilities in Faster ScoreCam. Explainable AI can detect the real and fake faces in this study which is our ultimate goal. It can be visualized what the model sees in terms of DeepFake Face Detection. 

\begin{table}[h]
    \caption{Performance Metrics of Weighted Average on FaceForensics++ Dataset
}
    \label{tab:pm-ff++}
    \centering
    \begin{tabular}{|l|c|c|c|c|}
        \hline
        \textbf{Model}     & \textbf{Accuracy} & \textbf{Precision} & \textbf{Recall} & \textbf{F1-Score} \\
        \hline
        InceptionResNetV2  & \textbf{94\%}     & 0.94              & 0.94            & 0.94 \\ \hline
        XceptionNet        & 93\%           & 0.93              & 0.93            & 0.93 \\ \hline
        EfficientNetV2S    & 92\%           & 0.92              & 0.92            & 0.92 \\ \hline
        EfficientNetV2M    & 88\%           & 0.89              & 0.88            & 0.88 \\ \hline
    \end{tabular}
\end{table}

Table \ref{tab:pm-ff++} shows that the models also had high levels of accuracy, with InceptionResNetV2 resulting with 94\%, XceptionNet coming in second with 93\%, and EfficientNetV2S coming in third with 92\%. It's important to note that these models did well overall on the FaceForensics++ Dataset, even though EfficientNetV2M had a slightly lower accuracy of 88\%. 

For the Faceforensics++ dataset, the Expainable AI has proven much more potential in the case of GradCam, GradCam++, and Faster ScoreCam. All the models for this dataset have more robustness than the CelebDF-V2 dataset which is shown in Figure \ref{fig:xai-xn-ff} to Figure \ref{fig:xai-env2-ff}. Explainable AI can also detect real and fake faces from the Faceforensics++ dataset. It can be visualized more accurately what the model sees in terms of DeepFake Face Detection in this particular case.

\subsection{Performance Comparison}
In this part, we compared our result with some of earlier research. With an excellent accuracy rate of 98\%, our proposed deepfake face identification algorithm exceeds earlier state-of-the-art techniques. The outcome proves how effective our strategy was. This accuracy, in particular, performs better than a number of important previous studies in the field. A Deep Neural Network architecture developed by Lee and Kim \cite{lee2021deepfake} obtained 97\% accuracy. In their study, Xu et al. \cite{xu2021deepfake} reached an accuracy of 91.2\%. While average and maximum pooling strategies were able to achieve an accuracy of 85.24\%, the proposed method of Kohli et al. \cite{kohli2021detecting} produced a maximum accuracy of 86.08\%. Two different models were introduced by Kim et al. \cite{kim2021fretal}, and their proposed model successfully detected deepfake videos with an accuracy of 86.97\%.


\section{Advantages and Limitations}
The deepfake video detection method used in this study relies on extracting key frames as a resource-effective strategy. Our research has shown that it is possible to build reliable models for deepfake video detection even with limited resources and its findings are encouraging. The interpretability of our models has been improved by our usage of Explainable AI techniques, offering important insights into their decision-making processes. The lack of high-quality video datasets for training and evaluation was one of the main issues we faced, which made it difficult to fully explore the potential of the models. However, more deepfake video datasets are expected to be released in the future, which will considerably increase the accuracy of our research and the overall efficiency of deepfake detection systems. Our models will have the chance to improve their comprehension of deepfake patterns and characteristics as these datasets become available. 

\section{Future Works}
We believe that Explainable AI applications will develop further, possibly producing even better outcomes in terms of model interpretability and decision explanation. This innovation will not only improve our current detection method but also increase our understanding of the methods behind deepfake video production and detection. Furthermore, the field of deepfake detection is dynamic, with continuous advancements in AI and machine learning. Future research in this field may present cutting-edge models and methodologies that are superior to the capabilities of existing systems, ultimately resulting in more dependable and robust deepfake detection systems. These developments will be necessary to maintain the integrity of online material and keep up with the deepfake technology field.


\bibliographystyle{IEEEtran}
\bibliography{references}

\begin{thebibliography}{10}
\providecommand{\url}[1]{#1}
\csname url@samestyle\endcsname
\providecommand{\newblock}{\relax}
\providecommand{\bibinfo}[2]{#2}
\providecommand{\BIBentrySTDinterwordspacing}{\spaceskip=0pt\relax}
\providecommand{\BIBentryALTinterwordstretchfactor}{4}
\providecommand{\BIBentryALTinterwordspacing}{\spaceskip=\fontdimen2\font plus
\BIBentryALTinterwordstretchfactor\fontdimen3\font minus \fontdimen4\font\relax}
\providecommand{\BIBforeignlanguage}[2]{{%
\expandafter\ifx\csname l@#1\endcsname\relax
\typeout{** WARNING: IEEEtran.bst: No hyphenation pattern has been}%
\typeout{** loaded for the language `#1'. Using the pattern for}%
\typeout{** the default language instead.}%
\else
\language=\csname l@#1\endcsname
\fi
#2}}
\providecommand{\BIBdecl}{\relax}
\BIBdecl

\bibitem{wubet2020deepfake}
W.~M. Wubet, ``The deepfake challenges and deepfake video detection,'' \emph{Int. J. Innov. Technol. Explor. Eng}, vol.~9, 2020.

\bibitem{mitra2021machine}
A.~Mitra, S.~P. Mohanty, P.~Corcoran, and E.~Kougianos, ``A machine learning based approach for deepfake detection in social media through key video frame extraction,'' \emph{SN Computer Science}, vol.~2, pp. 1--18, 2021.

\bibitem{tanvir2023explainable}
M.~Tanvir Rouf~Shawon, G.~Shahariar~Shibli, F.~Ahmed, and S.~K. Saha~Joy, ``Explainable cost-sensitive deep neural networks for brain tumor detection from brain mri images considering data imbalance,'' \emph{arXiv e-prints}, pp. arXiv--2308, 2023.

\bibitem{zhou2005training}
Z.-H. Zhou and X.-Y. Liu, ``Training cost-sensitive neural networks with methods addressing the class imbalance problem,'' \emph{IEEE Transactions on knowledge and data engineering}, vol.~18, no.~1, pp. 63--77, 2005.

\bibitem{lee2021deepfake}
G.~Lee and M.~Kim, ``Deepfake detection using the rate of change between frames based on computer vision,'' \emph{Sensors}, vol.~21, no.~21, p. 7367, 2021.

\bibitem{rossler2019faceforensics++}
A.~Rossler, D.~Cozzolino, L.~Verdoliva, C.~Riess, J.~Thies, and M.~Nie{\ss}ner, ``Faceforensics++: Learning to detect manipulated facial images,'' in \emph{Proceedings of the IEEE/CVF international conference on computer vision}, 2019, pp. 1--11.

\bibitem{dolhansky2020deepfake}
B.~Dolhansky, J.~Bitton, B.~Pflaum, J.~Lu, R.~Howes, M.~Wang, and C.~C. Ferrer, ``The deepfake detection challenge (dfdc) dataset,'' \emph{arXiv preprint arXiv:2006.07397}, 2020.

\bibitem{xu2021deepfake}
B.~Xu, J.~Liu, J.~Liang, W.~Lu, and Y.~Zhang, ``Deepfake videos detection based on texture features.'' \emph{Computers, Materials \& Continua}, vol.~68, no.~1, 2021.

\bibitem{korshunov2018deepfakes}
P.~Korshunov and S.~Marcel, ``Deepfakes: a new threat to face recognition? assessment and detection,'' \emph{arXiv preprint arXiv:1812.08685}, 2018.

\bibitem{li2019celeb}
Y.~Li, X.~Yang, P.~Sun, H.~Qi, and S.~Lyu, ``Celeb-df (v2): a new dataset for deepfake forensics [j],'' \emph{arXiv preprint arXiv}, 2019.

\bibitem{kohli2021detecting}
A.~Kohli and A.~Gupta, ``Detecting deepfake, faceswap and face2face facial forgeries using frequency cnn,'' \emph{Multimedia Tools and Applications}, vol.~80, pp. 18\,461--18\,478, 2021.

\bibitem{kim2021fretal}
M.~Kim, S.~Tariq, and S.~S. Woo, ``Fretal: Generalizing deepfake detection using knowledge distillation and representation learning,'' in \emph{Proceedings of the IEEE/CVF conference on computer vision and pattern recognition}, 2021, pp. 1001--1012.

\bibitem{Chollet_2017_CVPR}
F.~Chollet, ``Xception: Deep learning with depthwise separable convolutions,'' in \emph{Proceedings of the IEEE Conference on Computer Vision and Pattern Recognition (CVPR)}, July 2017.

\bibitem{tan2021efficientnetv2}
M.~Tan and Q.~V. Le, ``Efficientnetv2: Smaller models and faster training,'' 2021.

\bibitem{szegedy2016inceptionv4}
C.~Szegedy, S.~Ioffe, V.~Vanhoucke, and A.~Alemi, ``Inception-v4, inception-resnet and the impact of residual connections on learning,'' 2016.

\bibitem{gunning2019xai}
D.~Gunning, M.~Stefik, J.~Choi, T.~Miller, S.~Stumpf, and G.-Z. Yang, ``Xai—explainable artificial intelligence,'' \emph{Science robotics}, vol.~4, no.~37, p. eaay7120, 2019.

\end{thebibliography}
\vspace{12pt}
\color{red}

\end{document}